\documentclass[letterpaper]{article} % DO NOT CHANGE THIS
\usepackage{aaai2026}  % DO NOT CHANGE THIS
\usepackage{times}  % DO NOT CHANGE THIS
\usepackage{helvet}  % DO NOT CHANGE THIS
\usepackage{courier}  % DO NOT CHANGE THIS
\usepackage[hyphens]{url}  % DO NOT CHANGE THIS
\usepackage{graphicx} % DO NOT CHANGE THIS
\usepackage{natbib}  % DO NOT CHANGE THIS AND DO NOT ADD ANY OPTIONS TO IT
\usepackage{caption} % DO NOT CHANGE THIS AND DO NOT ADD ANY OPTIONS TO IT
\usepackage{algorithm}
\usepackage{algorithmic}
\usepackage{subcaption}

\usepackage{amsmath}
\usepackage{booktabs}
\usepackage{caption}
\usepackage{multicol}
\usepackage{multirow}
\usepackage{newfloat}
\usepackage{listings}
\DeclareCaptionStyle{ruled}{labelfont=normalfont,labelsep=colon,strut=off} % DO NOT CHANGE THIS
\floatstyle{ruled}
\newfloat{listing}{tb}{lst}{}
\floatname{listing}{Listing}
\title{LongT2IBench: A Benchmark for Evaluating Long Text-to-Image Generation with Graph-structured Annotations}
\author{
    Zhichao Yang\textsuperscript{\rm 1},
	Tianjiao Gu\textsuperscript{\rm 1},
	Jianjie Wang\textsuperscript{\rm 1},
	Feiyu Lin\textsuperscript{\rm 1},
	Xiangfei Sheng\textsuperscript{\rm 1},\\
    Pengfei Chen\textsuperscript{\rm 1*},
	Leida Li\textsuperscript{\rm 1, 2}\thanks{Corresponding authors.}
}
\affiliations{
	\textsuperscript{\rm 1}School of Artificial Intelligence, Xidian University, 
	\textsuperscript{\rm 2}State Key Laboratory of Electromechanical Integrated Manufacturing of High-Performance Electronic Equipments, Xidian University\\
	{(yangzhichao, gutianjiao, wangjianjie, linfeiyu)}@stu.xidian.edu.cn, xiangfeisheng@gmail.com, \\{(chenpengfei, ldli)}@xidian.edu.cn
}

\usepackage{bibentry}
\begin{document}

\maketitle

\begin{abstract}
	The increasing popularity of long Text-to-Image (T2I) generation has created an urgent need for automatic and interpretable models that can evaluate the image-text alignment in long prompt scenarios. However, the existing T2I alignment benchmarks predominantly focus on short prompt scenarios and only provide MOS or Likert scale annotations. This inherent limitation hinders the development of long T2I evaluators, particularly in terms of the interpretability of alignment. In this study, we contribute \textbf{LongT2IBench}, which comprises 14K long text-image pairs accompanied by graph-structured human annotations. Given the detail-intensive nature of long prompts, we first design a Generate-Refine-Qualify annotation protocol to convert them into textual graph structures that encompass entities, attributes, and relations. Through this transformation, fine-grained alignment annotations are achieved based on these granular elements. Finally, the graph-structed annotations are converted into alignment scores and interpretations to facilitate the design of T2I evaluation models. Based on LongT2IBench, we further propose \textbf{LongT2IExpert}, a LongT2I evaluator that enables multi-modal large language models (MLLMs) to provide both quantitative scores and structured interpretations through an instruction-tuning process with Hierarchical Alignment Chain-of-Thought (CoT). Extensive experiments and comparisons demonstrate the superiority of the proposed LongT2IExpert in alignment evaluation and interpretation. Data and code have been released in https://welldky.github.io/LongT2IBench-Homepage/.
\end{abstract}

% Uncomment the following to link to your code, datasets, an extended version or similar.
% You must keep this block between (not within) the abstract and the main body of the paper.
% \begin{links}
	%     \link{Code}{https://aaai.org/example/code}
	%     \link{Datasets}{https://aaai.org/example/datasets}
	%     \link{Extended version}{https://aaai.org/example/extended-version}
	% \end{links}

\begin{figure}[ht]
	\centering
	\includegraphics[width=0.97\columnwidth]{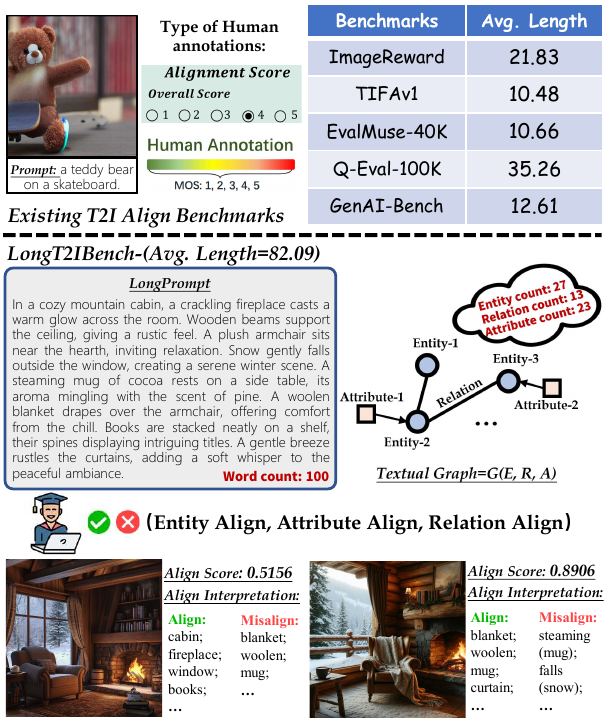} % Reduce the figure size so that it is slightly narrower than the column. Don't use precise values for figure width.This setup will avoid overfull boxes.
	\caption{Illustration of the graph-structured annotations of LongT2IBench. Compared to existing benchmarks, we focus on T2I alignment in long prompt scenarios, offering both quantitative scores and fine-grained interpretations.}
	\label{fig1}
\end{figure}

\section{Introduction}

With the extensive application of Text-to-Image (T2I) generation models in fields such as artistic creation and advertising design, user demand for long T2I generation capabilities has significantly increased \cite{ko2023large, zhao2024enhancing, 10168279, sheng2025agiqa, yang2025language}. This broadened research efforts from short prompts to increasingly complex and lengthy inputs \cite{hu2024ella, liu2024improving}. Long prompts inherently contain more elements and detailed descriptions, presenting significant challenges for alignment evaluation. Given the high costs and inefficiencies associated with manual evaluation, developing automatic and interpretable methods for evaluating long T2I alignment has become in urgent need. They can effectively evaluate the performance of existing T2I models and guide the optimization of future models. High-quality, fine-grained annotations of long T2I alignment serve as the essential foundation for developing such evaluation methods.

In the past few years, a proliferation of T2I alignment benchmarks have been proposed, along with corresponding evaluation algorithms. Early benchmarks, such as Pick-a-pic \cite{kirstain2023pick} and HPDv2 \cite{wu2023human} employed side-by-side comparisons to evaluate the alignment between generated image and text. Other benchmarks like ImageReward \cite{xu2023imagereward} and EvalAlign \cite{tan2024evalalign} utilized Likert scale annotation, with human annotators providing holistic scores based on general text-image correspondence. Recently, Benchmarks like T2I-CompBench \cite{huang2023t2i} and GenAI-Bench \cite{li2024genai} introduced curated prompts designed to evaluate specific capabilities of generative models, including counting, spatial relationships, etc. For fine-grained evaluation, benchmarks like TIFA \cite{hu2023tifa} and EvalMuse-40K \cite{han2024evalmuse} decomposed prompts into multiple concise phrases, formulating separate assessments that require human annotators to provide element-wise alignment scores.

While existing benchmarks provide valuable insights, they predominantly focus on short prompt scenarios and only provide Mean Opinion Score (MOS) or Likert-scale human annotations (Figure \ref{fig1}-top). This inherent limitation hinders the development of long T2I evaluators, particularly regarding alignment interpretability. The detail-intensive nature of long prompts presents two essential challenges for obtaining high-quality, fine-grained alignment annotations: (1) \textbf{Detail Overload}: it is both challenging and unreliable for annotators to directly provide an overall alignment score. (2) \textbf{Alignment Complexity}: element-wise annotations are inadequate in capturing the intricate alignment relationships present in long prompts, such as connections between distant elements. To address these challenges, we contribute \textbf{LongT2IBench}, a benchmark specifically designed for long T2I alignment that provides both quantitative alignment scores and fine-grained interpretations by introducing graph-structured annotations (Figure \ref{fig1}-bottom). LongT2IBench consists of 3K long prompts and 14K image-text pairs. To ensure the diversity, long prompts are obtained from three sources: human-generated content, AI-generated content, and long image captions. These prompts are first converted into textual graph structures that encompass entities, attributes, and relations. This conversion follows a Generate-Refine-Qualify protocol to ensure exceptional structural fidelity. Through this transformation, fine-grained alignment annotations are achieved based on these granular elements of textual graphs. We also implement a re-verification process to ensure the reliability of the alignment annotations. Finally, the above annotations are converted into alignment scores and interpretations.

Building upon LongT2IBench, we further propose \textbf{LongT2IExpert}, a long T2I evaluator which empowers multi-modal large language models (MLLMs) to provide both quantitative scores and structured interpretations. Inspired by the construction of LongT2IBench, we first introduce a Hierarchical Alignment Chain-of-Thought (HA-CoT) to prompt MLLMs to engage in structured alignment reasoning. HA-CoT features a three-hop reasoning process, progressively guiding models from fine-grained analysis to comprehensive evaluation. Then, human-annotated alignment scores and interpretations are utilized as supervision signals to fine-tune MLLMs in a multi-task manner. With these designs, we achieve a unified scoring and interpreting model that aligns closely with human evaluators.

The contributions of this work are three-fold:

\begin{itemize}
	\item We contribute LongT2IBench, a benchmark of Long Text-to-Image generation that contains 14K long text-image pairs with graph-structured annotations. It provides both quantitative scores and fine-grained interpretations, elucidating the intricate alignment relationships between textual elements and visual components.
	\item We propose LongT2IExpert, a novel long T2I evaluator that equips multimodal large language models (MLLMs) with structured alignment reasoning capabilities, simultaneously performing alignment scoring and interpreting.
	\item We establish a systematic framework for examining long T2I alignment abilities. It comprehensively assesses both scoring accuracy and interpretation quality, while thoroughly testing performance across varying text lengths and alignment dimensions: entity, attribute, and relation.
\end{itemize}

\section{Related Works}

\textbf{Text-to-Image Alignment Benchmarks.} Many benchmarks have been developed to evaluate the text-to-image alignment of generation models. Early benchmarks such as Pick-a-pic \cite{kirstain2023pick} and HPDv2 \cite{wu2023human}, compare models side-by-side to evaluate generated image quality. Other benchmarks like ImageReward \cite{xu2023imagereward} and EvalAlign \cite{tan2024evalalign} utilized Likert scale annotation to collect overall alignment scores. For fine-grained evaluation, benchmarks like TIFA \cite{hu2023tifa} and EvalMuse-40K \cite{han2024evalmuse} extract concise phrases from prompts and collect element-wise annotations. With increasing prompt length, benchmarks specifically designed for long T2I alignment remain scarce. In this paper, we contribute LongT2IBench, providing a foundation for examining and developing long T2I alignment evaluation models.

\textbf{Text-to-Image Alignment Evaluators.} Early T2I evaluators directly utilized vision-language models (VLMs) for alignment evaluation, such as CLIPScore \cite{hessel2021clipscore} and BLIP2Score \cite{li2023blip}. With the extensive collection of alignment annotations, PickScore \cite{kirstain2023pick} and ImageReward \cite{xu2023imagereward} fine-tuned VLMs to align with human ratings. With MLLMs rapidly advanced, VQAScore \cite{li2024evaluating} and Q-Eval-Score \cite{zhang2025q} leverage their powerful vision-language capabilities for T2I alignment. While existing models have achieved remarkable progress, their evaluation performance in long-prompt contexts remains largely unexplored. In this paper, we address this gap and introduce LongT2IExpert, a model specifically designed for long T2I alignment.

\begin{figure*}[t]
	\centering
	\includegraphics[width=0.99\textwidth]  {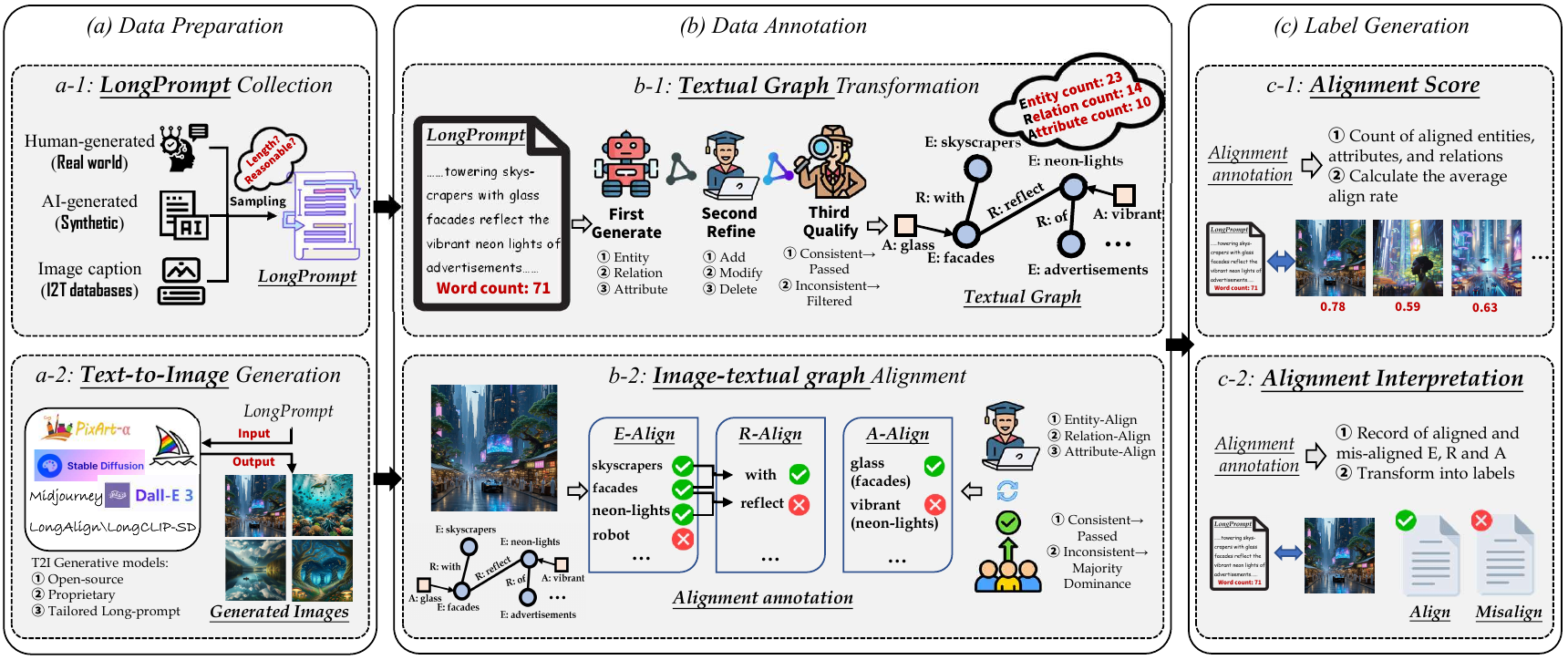}
	\caption{Overview of the construction pipeline for \textbf{LongT2IBench}. The pipeline consists of three stages: (a) \textbf{Data Preparation.} Long prompts are collected from three sources and input into various T2I models to generate images. (b) \textbf{Data Annotation.} Long prompts are converted into textual graph structures, and fine-grained image-textual graph alignment annotations are achieved. (c) \textbf{Label Generation.} Two categories of labels: quantitative alignment scores and alignment interpretations are produced based on graph-structured human annotations.}
	\label{fig2}
\end{figure*}

\textbf{Text-to-Image Models for Long Prompt.} Traditional CLIP-based T2I generative models face limitations of token length restrictions, hindering their ability to process long prompts \cite{radford2021learning}. To address the dilemma, recent researches employed large language models (LLMs) as text encoders to improve semantic extraction, with representative models such as Stable Diffusion 3.5 \cite{esser2024scaling}. Alternatively, other approaches introduced architectural improvements to the CLIP to accommodate longer inputs. LongCLIP \cite{ zhang2024long} offers a plug-and-play alternative to standard CLIP that supports long-text processing. LongAlign \cite{liu2024improving} decomposes long prompts into individual sentences for separate encoding, enabling long T2I generation. The rapid advancement in long-prompt generation has created an urgent need for automatic, interpretable long T2I alignment evaluation models.

\section{LongT2IBench}

In this section, we detail LongT2IBench, a benchmark for evaluating long text-to-image alignment by introducing graph-structured annotations. The overview of construction pipeline for LongT2IBench is illustrated in Figure \ref{fig2}. Throughout the benchmark, we employ multiple strategies to ensure both diversity and reliability. Next, we will elaborate on the construction process from three aspects: data preparation, data annotation, and label generation, followed by a statistical analysis.

\subsection{Data Preparation}

\textbf{Long Prompt Collection.} To ensure prompt diversity and comprehensive evaluation of long T2I alignment capabilities, we collect long prompts from three sources and controlled length distribution across different word-count intervals. Specifically, the long prompts are sampled from human-generated content (Human-Gen), AI-generated content (AI-Gen), and long image captions (Img-Cap). Human-Gen prompts are sourced from DiffusionDB \cite{wang2023diffusiondb}, which contains numerous prompts created by real users. AI-Gen prompts are synthesized using GPT-4 \cite{achiam2023gpt} following specific templates. Img-Cap prompts are derived from DOCCI \cite{onoe2024docci}, where each image is accompanied by a long description with fine-grained visual details. Meanwhile, we ensured balanced sampling across five word-count intervals (30-50, 50-70, 70-90, 90-110, 110+) to systematically assess T2I evaluators' alignment capabilities at varying text lengths.

\textbf{Text-to-Image Generation.} To ensure the diversity of the generated images, we select six distinct types of text-to-image generative models. These models are categorized into three groups: (1) basic open-source popular T2I models, namely Stable Diffusion v3.5 \cite{esser2024scaling} and PixArt-$\alpha$ \cite{chenpixart}; (2) powerful proprietary T2I models, including DALL-E 3 \cite{dalle} and Midjourney v6 \cite{midjourney}; (3) tailored long T2I models, including LongCLIP-SD \cite{zhang2024long} and LongSD \cite{liu2024improving}. For each long prompt, we utilize the default settings of these models to generate images, ensuring a comprehensive dataset for annotation.

\subsection{Data Annotation}

\textbf{Textual Graph Transformation.} Graph representation has proven a powerful paradigm for modeling complex relationships and structural information in various domains \cite{hamilton2020graph, yang2024semantics}. Given the detail-intensive nature of long prompts, we perform textual graph transformation on the collected prompts. This establishes the foundation for more objective and accurate text-image alignment annotations. Additionally, the graph structure enables precise localization of aligned and misaligned elements, allowing us to better evaluate long T2I alignment capabilities.

\begin{figure}[t]
	\centering
	\includegraphics[width=0.9\columnwidth]{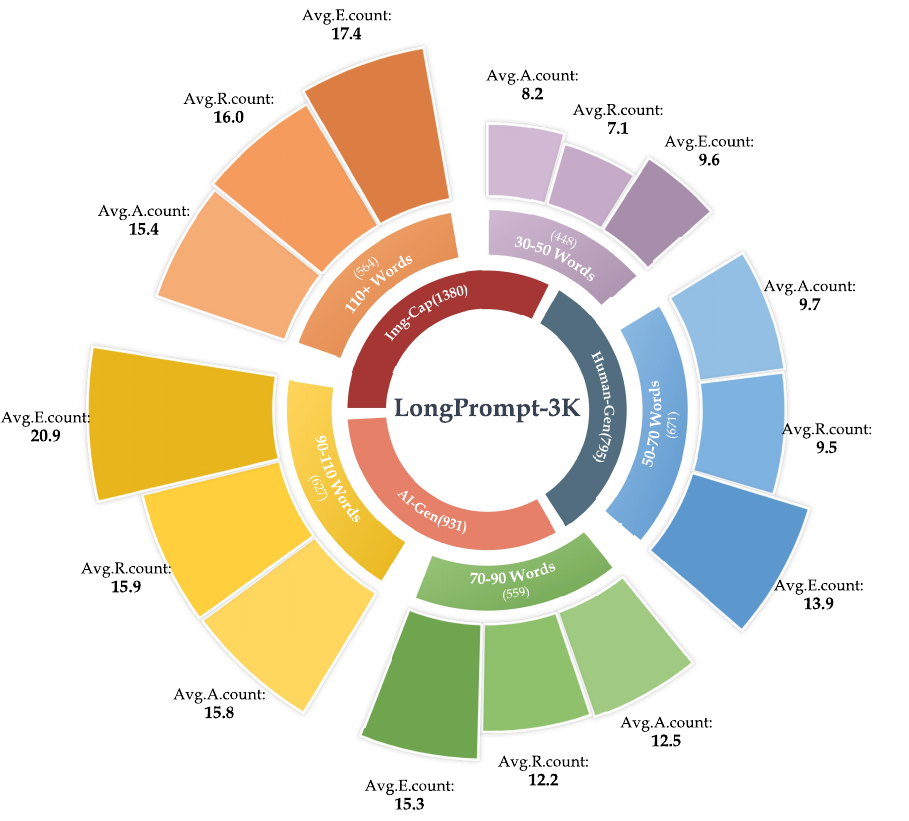} % Reduce the figure size so that it is slightly narrower than the column. Don't use precise values for figure width.This setup will avoid overfull boxes.
	\caption{The Source Distribution Map of \textbf{LongPrompt-3K}. The long prompts are sampled from three sources: Human-Gen, AI-Gen and Img-Cap. These prompts are evenly distributed across different word count ranges, with the number of entities, attributes, and relationships in each range also being statistically presented. }
	\label{fig3}
\end{figure}

To ensure exceptional graph structural fidelity, we adopt a systematic three-phase transformation protocol. First, we design specific input prompts and output formats to guide GPT-4 to \textbf{generate} textual graphs, converting long prompts into graph-structures comprising entities, attributes, and relations. Second, trained human annotators \textbf{refine} these generated graph structures to ensure accuracy. This refinement process involves removing inappropriate conversions and making detailed adjustments through additions, deletions, and modifications to coarse transformations. Third, we perform a double-check to \textbf{qualify} the refined transformations from the previous stage, ensuring reliability. Only conversion graphs that receive consistent agreement are retained in the final dataset. This rigorous multi-stage process ensures high-quality structural graph transformation that accurately capture the semantic relationships of long prompts. Throughout this process, 3K precise conversions were retained from an initial collection of 4.5K long prompts. A statistical analysis on the number of entities, attributes, and relations within the transformed graph structures in \textbf{LongPrompt-3K} are illustrated in Figure \ref{fig3}. The instructions for graph generation using GPT-4, the operational interface for Refine and Qualify stages, along with annotations training guidelines are detailed in the Appendix.

\textbf{Image-textual graph Alignment.} Based on transformed textual graphs, we implement an initial annotation followed by a re-verification process to obtain reliable image-textual graph alignment annotations. Trained annotators perform initial annotations, making binary decisions on the alignment between transformed entities, attributes, and relations with the corresponding images, namely E-Align, A-Align, and R-Align respectively. To improve efficiency, we incorporate hierarchical annotation logic into our platform design, where annotators first evaluate entities, and the platform automatically filters out attributes and relationships associated with misaligned entities. Furthermore, we engage three independent annotators to review the initial annotation results, removing instances with significant disagreement while retaining alignment annotations that achieve majority consensus. After annotation, 14K image-text pairs are retained from the original 18K samples for the final dataset. Discarded data includes images containing NSFW content, as well as instances with severe generation distortions.

\begin{figure}[t]
	\centering
	\begin{subfigure}[b]{0.49\columnwidth}
		\centering
		\includegraphics[width=\textwidth]{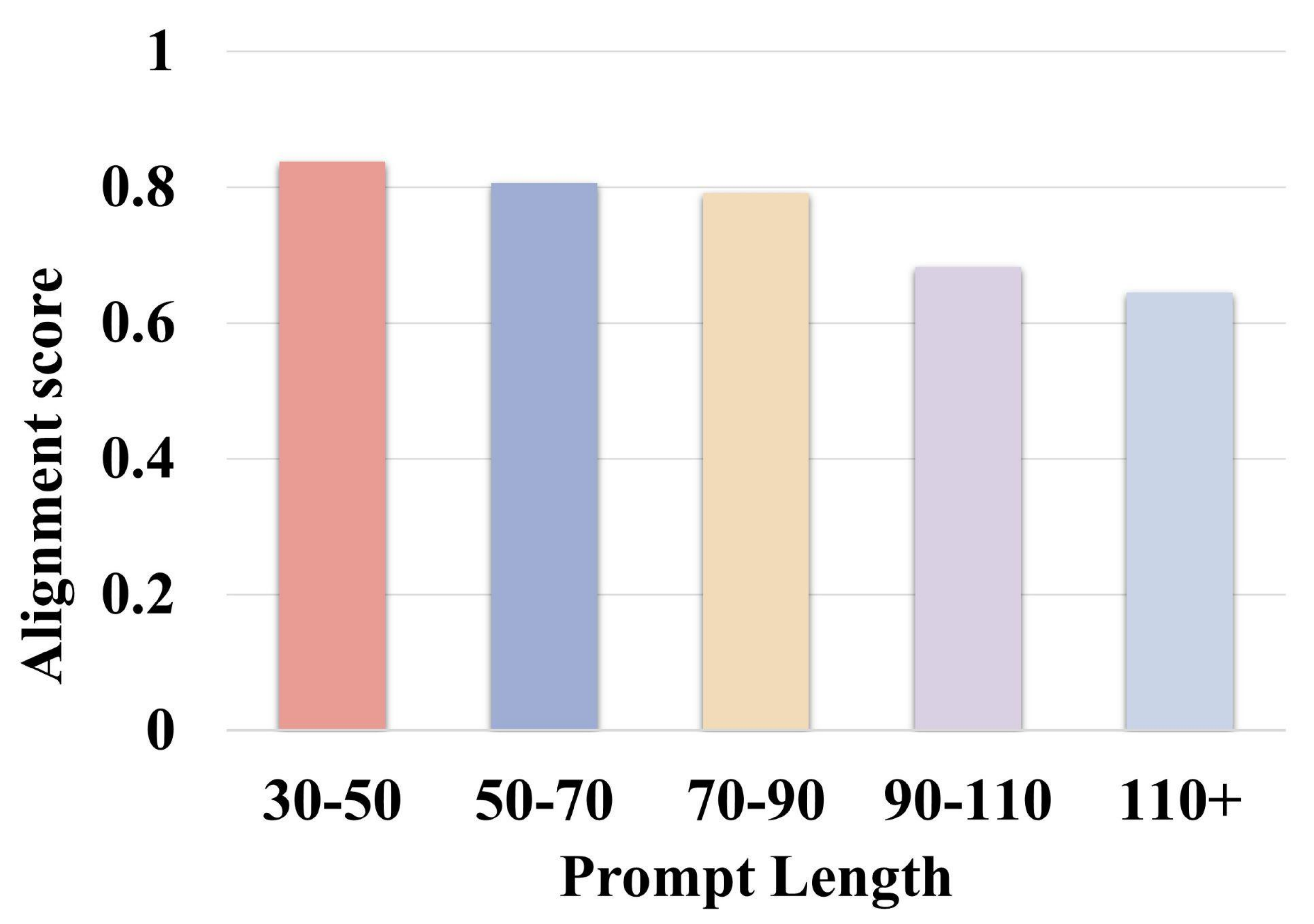}
		\caption{}
		\label{fig:image1}
	\end{subfigure}%
	\hfill
	\begin{subfigure}[b]{0.49\columnwidth}
		\centering
		\includegraphics[width=\textwidth]{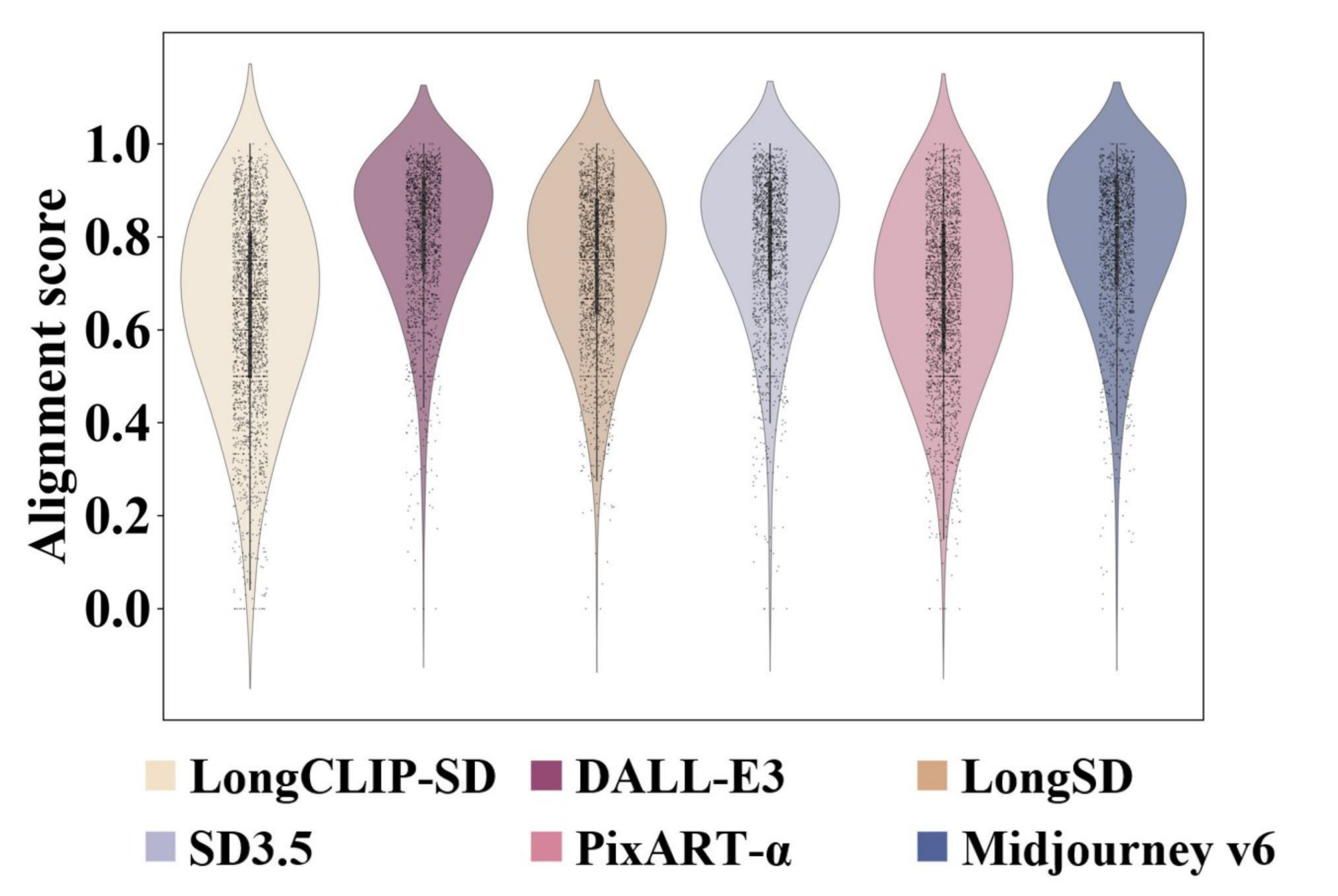}
		\caption{}
		\label{fig:image2}
	\end{subfigure}
	
	%	\vspace{0.3cm}
	
	\begin{subfigure}[b]{0.49\columnwidth}
		\centering
		\includegraphics[width=\textwidth]{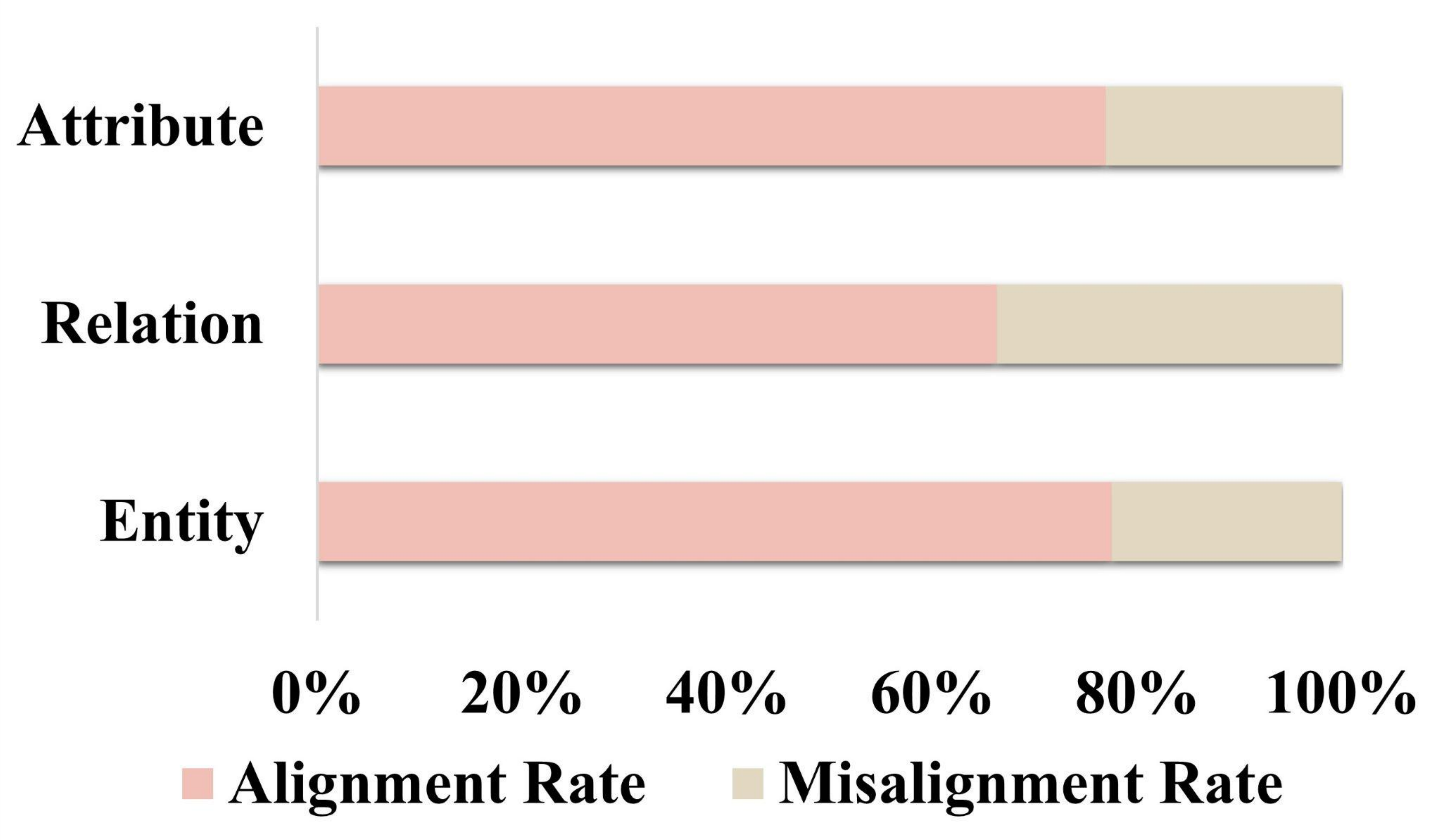}
		\caption{}
		\label{fig:image3}
	\end{subfigure}%
	\hfill
	\begin{subfigure}[b]{0.49\columnwidth}
		\centering
		\includegraphics[width=\textwidth]{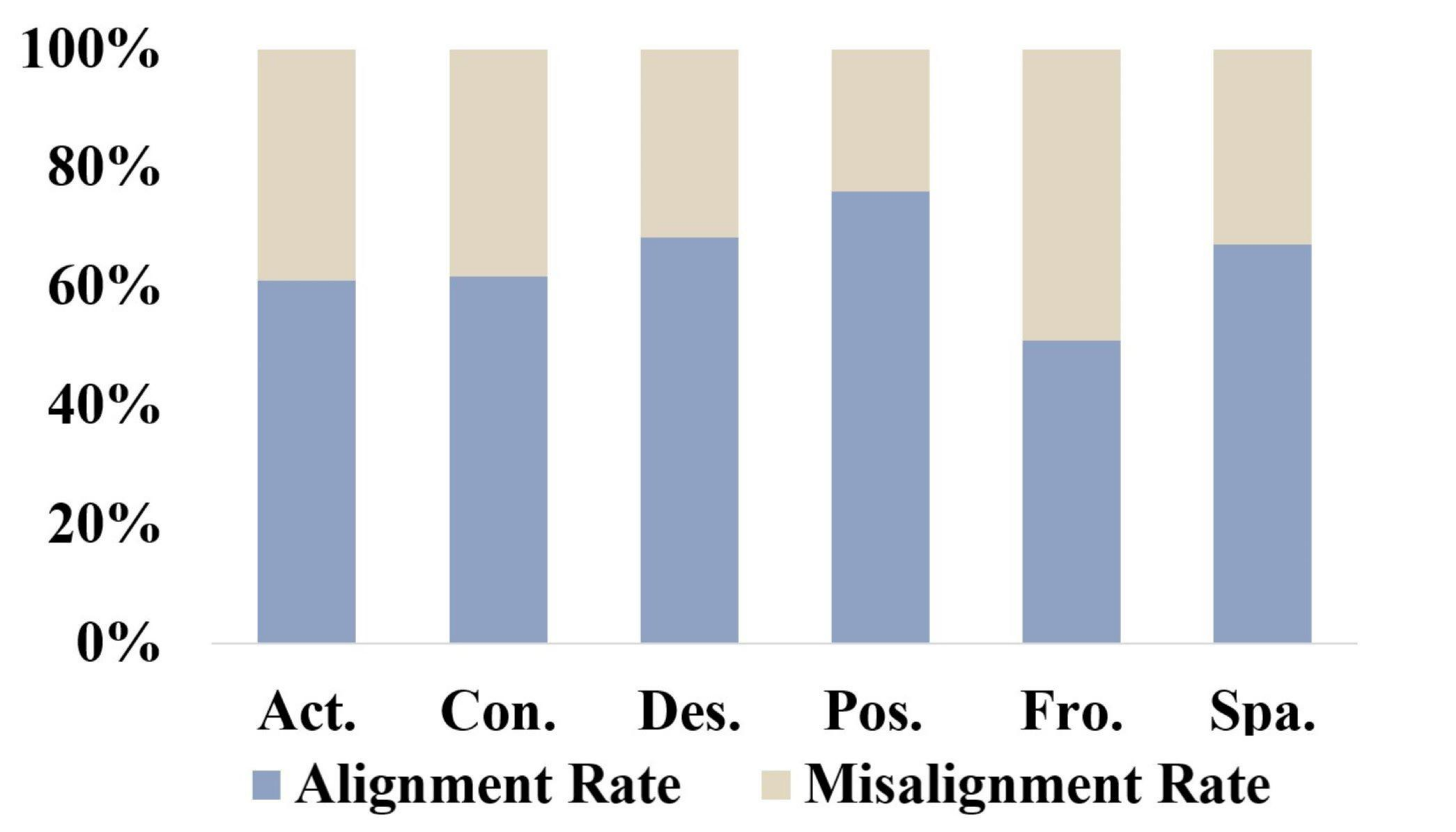}
		\caption{}
		\label{fig:image4}
	\end{subfigure}
	\caption{Statistical Analysis. (a) Average alignment scores across five word-count intervals. (b) Distribution of annotated alignment scores for six T2I generative models. (c) Alignment and misalignment rates across entities, attributes, and relations. (d) Alignment percentages among six relation categories (Action, Connection, Description, Possession, From/to and Spatial relation).}
	\label{fig4}
\end{figure}

\subsection{Label Generation}

\textbf{Alignment Score.} We envision future text-to-image generative models capable of aligning with all textual details. Therefore, we treat all entities, attributes, and relations in the textual graph with equal importance. The alignment score is defined as the ratio of aligned elements to the total number. We analyzed the average alignment scores across five word-count intervals, as illustrated in Figure \ref{fig4}(a). The statistical results demonstrate that alignment scores gradually decrease as prompt length increases, underscoring the importance of evaluating text-image alignment in long-prompt scenarios. Additionally, Figure \ref{fig4}(b) shows the distribution of alignment annotations across six generative models, revealing performance disparities in long-text alignment capabilities, with proprietary models (DALL-E 3 and Midjourney v6) demonstrating relative advantages.

\textbf{Alignment Interpretation.} In long-prompt scenarios, the need for alignment interpretation is more critical, particularly in localizing misaligned elements that can guide optimization for generation models. Leveraging textual graph transformation and image-textual graph alignment annotations, we can derive structured alignment interpretations for each image-text pair, represented as aligned and misaligned entities, attributes, and relations. We analyzed the proportions of alignment and misalignment across these three categories, as illustrated in Figure \ref{fig4}(c). The results reveal that relation misalignment constitutes a particularly significant proportion. During the annotation process, we instructed annotators to classify relations into distinct categories, with Figure \ref{fig4}(d) presenting the alignment and misalignment proportions across these relation types.

\begin{figure}[t]
	\centering
	\includegraphics[width=\columnwidth]{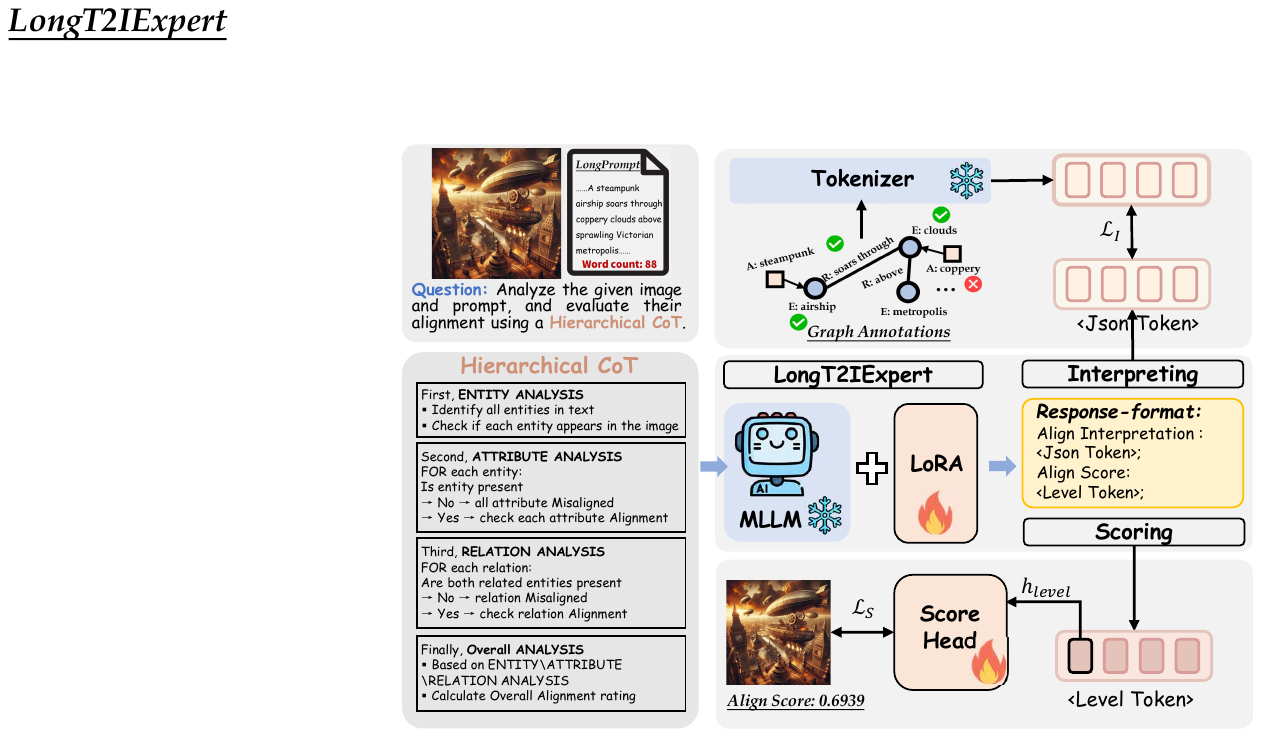} % Reduce the figure size so that it is slightly narrower than the column. Don't use precise values for figure width.This setup will avoid overfull boxes.
	\caption{Overall pipeline of the proposed \textbf{LongT2IExpert}. A Hierarchical Alignment Chain-of-Thought (CoT) is designed to instruct MLLMs for structured alignment reasoning. Numerical alignment scores and graph-structured interpretations are utilized to train MLLMs for alignment scoring and interpreting in a multi-task manner.}
	\label{fig5}
\end{figure}

\section{LongT2IExpert}

Building on the proposed LongT2IBench, we propose \textbf{LongT2IExpert}, a long T2I evaluator that enables multi-modal large language models (MLLMs) to provide both quantitative scores and structured interpretations through an instruction-tuning process with Hierarchical Alignment Chain-of-Thought (HA-CoT). The overall pipeline is illustrated in Figure \ref{fig5}.

\subsection{Instructing MLLMs for Alignment Reasoning}

The evaluation of Long T2I alignment requires that models possess both the capability to analyze long texts and the ability to achieve multimodal alignment. This imposes high demands on the foundational capabilities of the model, leading us to select a multimodal large language model as our backbone, i.e. Qwen2.5-VL-7B-Instruct \cite{bai2023qwen}. However, even with this advanced architecture, the characteristics of Detail Overload and Alignment Complexity inherent in this task render alignment evaluation a challenging endeavor. Inspired by the construction process of LongT2IBench, we introduce a Hierarchical Alignment Chain-of-Thought (HA-CoT) to guide MLLMs in mimicking human-like structured alignment reasoning.

HA-CoT features a three-hop reasoning process. In the initial stage, the MLLM acts as an Entity Aligner, analyzing all entities mentioned in the long text and determining their alignment with the image. In the second phase, acting as an Attribute and Relation Aligner, it examines the alignment of attributes and relations associated with each aligned entity. Finally, based on the previous analysis, it synthesizes the overall alignment situation and provides a comprehensive alignment score. Throughout the structured CoT reasoning process, MLLMs perform a multi-faceted alignment evaluation that mimics the deliberation of a human evaluator.

\subsection{Training MLLMs for Alignment Evaluation}

Building on the structured alignment reasoning process, we utilize high-quality alignment scores and interpretations as supervision signals for training MLLMs in alignment evaluation. Inspired by LISA \cite{lai2024lisa, sheng2025instructcrop}, which uses special tokens to enable reasoning segmentation capabilities in MLLMs, we expand the original LLM vocabulary with a new token $<$Level$>$ to signify the request for a numerical score output. Simultaneously, we convert the graph-based human annotations into JSON formats, preserving the original structured information while facilitating easier parsing by MLLMs. These approaches allow us to leverage the powerful understanding and reasoning capabilities of MLLMs for simultaneous alignment scoring and interpreting. Give a long prompt $x_{txt}$ along with the generated image $ x_{img}$ and pre-defined CoT, we feed them into the MLLM $\mathcal{F}$, which in turn outputs a text response $\hat{y} _{txt}$:
\begin{equation}
	\hat{y} _{txt} = \mathcal{F}\left ( x_{txt}, x_{img}, \text{CoT} \right ),
\end{equation}
where the output $\hat{y} _{txt}$ would include a $<$Json$>$ token and $<$Level$>$ token corresponding to the predicted alignment interpretation and alignment score, respectively.

We then extract the MLLM last-layer embedding $ \hat{\mathbf{h}} _{Level}$ corresponding to the $<$Level$>$ token and apply a score head $\mathcal{R}$ to obtain the predicted alignment score $\hat{y} _{s}$. The score loss $\mathcal{L}_{S}$ is calculated using Mean Squared Error (MSE) between the predicted score and the human-annotated score $y _{s}$:
\begin{equation}
	\hat{y}_{s} = \mathcal{R}\left(\hat{\mathbf{h}}_{Level}\right), \mathcal{L}_{S} = (\hat{y}_{s} - y_{s})^2.
\end{equation}
For alignment interpretation, we decode the $<$Json$>$ token into the graph-structed interpretation $ \hat{\mathbf{G}}\left ( E,R,A \right )$, where $E, R, A$ represent entities, relations, and attributes, respectively. The interpretation loss $\mathcal{L}_{I}$ is calculated using cross-entropy between the predicted interpretation and the ground-truth interpretation $\mathbf{G}$:
\begin{equation}
	\mathcal{L}_{I} = \mathbf{CE}\left(\hat{\mathbf{G}}, \mathbf{G}\right).
\end{equation}
We incorporate a parameter-efficient LoRA module \cite{hu2022lora} for fine-tuning the MLLM. The model is trained end-to-end using the score prediction loss $\mathcal{L}_{S}$ and the interpretations generation loss $\mathcal{L}_{I}$. The overall training objective $\mathcal{L}$ is the weighted sum of these losses:
\begin{equation}
	\mathcal{L} =\mathcal{L}_{I} + \lambda \cdot \mathcal{L}_{S},
\end{equation}
where the hyperparameter $\lambda$ balances the two losses.

\begin{table*}[t]
	\centering
	
	% --- Formatting Adjustments to fit the table within page margins without resizebox ---
	\small % Use 9pt font size, as allowed by academic guidelines for large tables.
	\setlength{\tabcolsep}{1mm} % Reduce space between columns (default is 6pt).
	
	\begin{tabular}{l|ll|ll|ll|ll|ll|lll}
		\toprule
		% --- Header Row 1: Using abbreviated headers ---
		\multirow{2}{*}[-0.65ex]{\textbf{Model}} & \multicolumn{2}{c|}{\textbf{30-50 words}} & \multicolumn{2}{c|}{\textbf{50-70 words}} & \multicolumn{2}{c|}{\textbf{70-90 words}} & \multicolumn{2}{c|}{\textbf{90-110 words}} & \multicolumn{2}{c|}{\textbf{110+ words}} & \multicolumn{3}{c}{\textbf{Overall}} \\
		\cmidrule(lr){2-3} \cmidrule(lr){4-5} \cmidrule(lr){6-7} \cmidrule(lr){8-9} \cmidrule(lr){10-11} \cmidrule(lr){12-14}
		
		% --- Header Row 2: Sub-metrics ---
		& \textbf{SRCC} & \textbf{PLCC} & \textbf{SRCC} & \textbf{PLCC} & \textbf{SRCC} & \textbf{PLCC} & \textbf{SRCC} & \textbf{PLCC} & \textbf{SRCC} & \textbf{PLCC} & \textbf{SRCC} & \textbf{PLCC} & \textbf{Avg.} \\
		\midrule 
		
		% --- Data Rows with Sub-optimal results underlined ---
		CLIPScore     & 0.224 & 0.235 & 0.349 & 0.370 & 0.271 & 0.240 & 0.208 & 0.202 & 0.209 & 0.184 & 0.269 & 0.264 & 0.267 \\
		BLIP2Score    & 0.384 & 0.231 & 0.342 & 0.358 & 0.190 & 0.155 & 0.215 & 0.191 & 0.149 & 0.061 & 0.266 & 0.223 & 0.245 \\
		PickScore     & 0.437 & 0.550 & 0.368 & 0.359 & 0.313 & 0.328 & 0.352 & 0.373 & 0.153 & 0.165 & 0.327 & 0.344 & 0.336 \\
		ImageReward   & 0.283 & 0.339 & 0.270 & 0.319 & 0.181 & 0.226 & 0.187 & 0.188 & -0.023 & 0.072 & 0.210 & 0.244 & 0.227 \\
		HPSv2         & 0.540 & 0.534 & 0.479 & 0.473 & \underline{0.394} & 0.417 & 0.395 & \underline{0.408} & 0.148 & 0.151 & 0.381 & 0.392 & 0.387 \\
		VQAScore      & 0.504 & 0.385 & 0.333 & 0.360 & 0.270 & 0.389 & 0.195 & 0.200 & 0.242 & 0.225 & 0.298 & 0.291 & 0.295 \\
		FGA-BLIP2     & 0.414 & 0.399 & 0.308 & 0.330 & 0.193 & 0.299 & 0.160 & 0.171 & 0.130 & 0.155 & 0.230 & 0.262 & 0.246 \\
		Q-Eval-Score  & 0.470 & 0.416 & 0.460 & 0.459 & 0.339 & 0.382 & 0.320 & 0.301 & \underline{0.422} & \underline{0.365} & 0.361 & 0.355 & 0.358 \\
		\midrule
		PickScore*   & 0.387 & 0.510 & 0.530 & 0.521 & 0.393 & 0.432 & 0.419 & 0.372 & 0.250 & 0.264 & \underline{0.438} & 0.439 & 0.438 \\
		ImageReward* & 0.538 & \underline{0.584} & \underline{0.546} & \underline{0.574} & 0.384 & 0.398 & \underline{0.399} & 0.396 & 0.167 & 0.195 & \underline{0.438} & 0.439 & \underline{0.439} \\
		FGA-BLIP2*   & \underline{0.570} & 0.568 & 0.514 & 0.527 & 0.371 & \underline{0.475} & 0.327 & 0.355 & 0.126 & 0.144 & 0.406 & \underline{0.444} & 0.425 \\
		\midrule
		% --- Final row with our model, bolded for emphasis ---
		\textbf{LongT2IExpert} & \textbf{0.781} & \textbf{0.789} & \textbf{0.605} & \textbf{0.626} & \textbf{0.548} & \textbf{0.558} & \textbf{0.435} & \textbf{0.443} & \textbf{0.431} & \textbf{0.375} & \textbf{0.558} & \textbf{0.557} & \textbf{0.557} \\
		\bottomrule
	\end{tabular}
	\caption{Performance comparison between the proposed LongT2IExpert and the state-of-art methods for \textbf{Long T2I Alignment Scoring}. Here, `*' denotes the fine-tuned versions of corresponding methods using the training data from LongT2IBench. The SRCC and PLCC across five word-count intervals are reported. Best in \textbf{bold}, second \underline{underlined}.}  
	\label{tab:1}
\end{table*}

\begin{table*}[t]
	\centering
	
	% --- Formatting Adjustments to fit the table within page margins without resizebox ---
	\small % Use 9pt font size, as allowed by academic guidelines for large tables.
	\setlength{\tabcolsep}{1mm} % Reduce space between columns (default is 6pt).
	
	\begin{tabular}{l|lll|lll|lll|lll}
		\toprule
		% --- Header Row 1: Main Categories ---
		\multirow{2}{*}[-0.7ex]{\textbf{Model}} & \multicolumn{3}{c|}{\textbf{Entity}} & \multicolumn{3}{c|}{\textbf{Attribute}} & \multicolumn{3}{c|}{\textbf{Relation}} & \multicolumn{3}{c}{\textbf{All}} \\
		\cmidrule(lr){2-4} \cmidrule(lr){5-7} \cmidrule(lr){8-10} \cmidrule(lr){11-13}
		
		% --- Header Row 2: Specific Accuracy metrics ---
		& \textbf{Align} & \textbf{Mis.} & \textbf{Overall} & \textbf{Align} & \textbf{Mis.} & \textbf{Overall} & \textbf{Align} & \textbf{Mis.} & \textbf{Overall} & \textbf{Align} & \textbf{Mis.} & \textbf{Overall} \\
		\midrule
		
		% --- Data Rows: All numerical values are converted to percentages with one decimal place ---
		Gemini-1.5-flash & 48.4\% & 23.2\% & 43.6\% & 26.4\% & 9.2\% & 23.2\% & 6.4\% & 3.2\% & 5.6\% & 29.7\% & 11.8\% & 25.9\% \\
		Gemini-1.5-pro   & 49.2\% & 36.1\% & \underline{46.7\%} & \underline{31.6\%} & \underline{16.6\%} & \underline{28.8\%} & \underline{13.6\%} & \underline{12.8\%} & \underline{13.4\%} & \underline{33.6\%} & 21.9\% & \underline{31.1\%} \\
		GPT-4o           & 44.2\% & 30.7\% & 41.6\% & 27.6\% & 13.9\% & 25.0\% & 10.0\% & 11.0\% & 10.2\% & 29.3\% & 18.6\% & 27.0\% \\
		GPT-4-Turbo      & 43.1\% & 27.1\% & 40.1\% & 24.9\% & 11.9\% & 22.5\% & 5.6\% & 5.0\% & 5.4\% & 26.8\% & 14.6\% & 24.2\% \\
		Grok-3           & 42.9\% & \textbf{44.0\%} & 43.1\% & 23.0\% & 16.4\% & 21.7\% & 6.2\% & 10.7\% & 7.4\% & 26.3\% & \underline{23.8\%} & 25.7\% \\
		\midrule
		
		Qwen2.5-v1-max   & \underline{49.9\%} & 22.8\% & 44.8\% & 28.2\% & 10.8\% & 24.9\% & 9.8\% & 4.7\% & 8.4\% & 31.7\% & 12.6\% & 27.7\% \\
		LLava-v1.6       & 33.6\% & 6.5\% & 28.4\% & 11.4\% & 1.9\% & 9.6\% & 1.9\% & 0.3\% & 1.5\% & 17.6\% & 2.9\% & 14.5\% \\
		\midrule
		
		% --- Data Row: "ours" model results, in bold ---
		\textbf{LongT2IExpert} & \textbf{80.2\%} & \underline{36.5\%} & \textbf{71.9\%} & \textbf{53.8\%} & \textbf{18.9\%} & \textbf{47.3\%} & \textbf{40.6\%} & \textbf{20.8\%} & \textbf{35.2\%} & \textbf{60.7\%} & \textbf{25.8\%} & \textbf{53.2\%} \\
		\bottomrule
	\end{tabular}
	\caption{Comparison between the proposed LongT2IExpert and the advanced multimodal large language models (MLLMs) for \textbf{Long T2I Alignment Interpreting}. Aligned and Misaligned accuracy across entity, attribute and relation are reported.}
	\label{tab:2}
\end{table*}

\section{Experiments}

\subsection{Experimental Settings}

\textbf{Implementation Details.} For LongT2IExpert training, we employ an open-source MLLM,  Qwen2.5-VL-7B-Instruct \cite{bai2023qwen}, and attach LoRA adapters (r=32, $\alpha$=64, dropout=0.05) for fine-tuning. The score head consists of three Linear layers. The learning rate is set to $5e^{-5}$ for LoRA parameters and $2e^{-4}$ for the score head. The hyperparameter $\lambda$ is set to 10, and the model is trained for 3 epochs on an A800 GPU. Details regarding LongT2IBench, including the annotation platforms and MLLM instructions for long prompt synthesis and textual graph transformations, as well as annotator information, are provided in the Appendix.

\textbf{Evaluation Setting.} We employ a standard train-validation-test split on LongT2IBench, ensuring long prompts are evenly distributed across five word-count intervals. For alignment score, we compare LongT2IExpert against two categories of existing state-of-the-art T2I evaluators. The first category comprises VLM-based models: CLIPScore \cite{hessel2021clipscore}, BLIP2Score \cite{li2023blip}, PickScore \cite{li2023blip}, ImageReward \cite{xu2023imagereward}, and FGA-BLIP2 \cite{han2024evalmuse}. The second category encompasses MLLM-based models: HPSv2 \cite{wu2023human}, VQAScore \cite{li2024evaluating}, and Q-Eval-Score \cite{zhang2025q}. To measure correlation with human annotations, we report both SRCC and PLCC metrics. Few models can simultaneously predict both alignment score and interpretations. To evaluate interpretation quality, we benchmark LongT2IExpert against two categories of MLLMs. Proprietary models include Gemini-1.5-flash/pro \cite{team2023gemini}, GPT-4o, GPT-4-Turbo \cite{achiam2023gpt}, and Grok-3 \cite{Grok}, while open-source models include Qwen2.5-vl-max \cite{bai2023qwen}, and LLava-v1.6 \cite{liu2023visual}. We use Accuracy to evaluate interpretation quality in identifying aligned/misaligned elements.

\begin{figure*}[t]
	\centering
	\includegraphics[width=0.92\textwidth]  {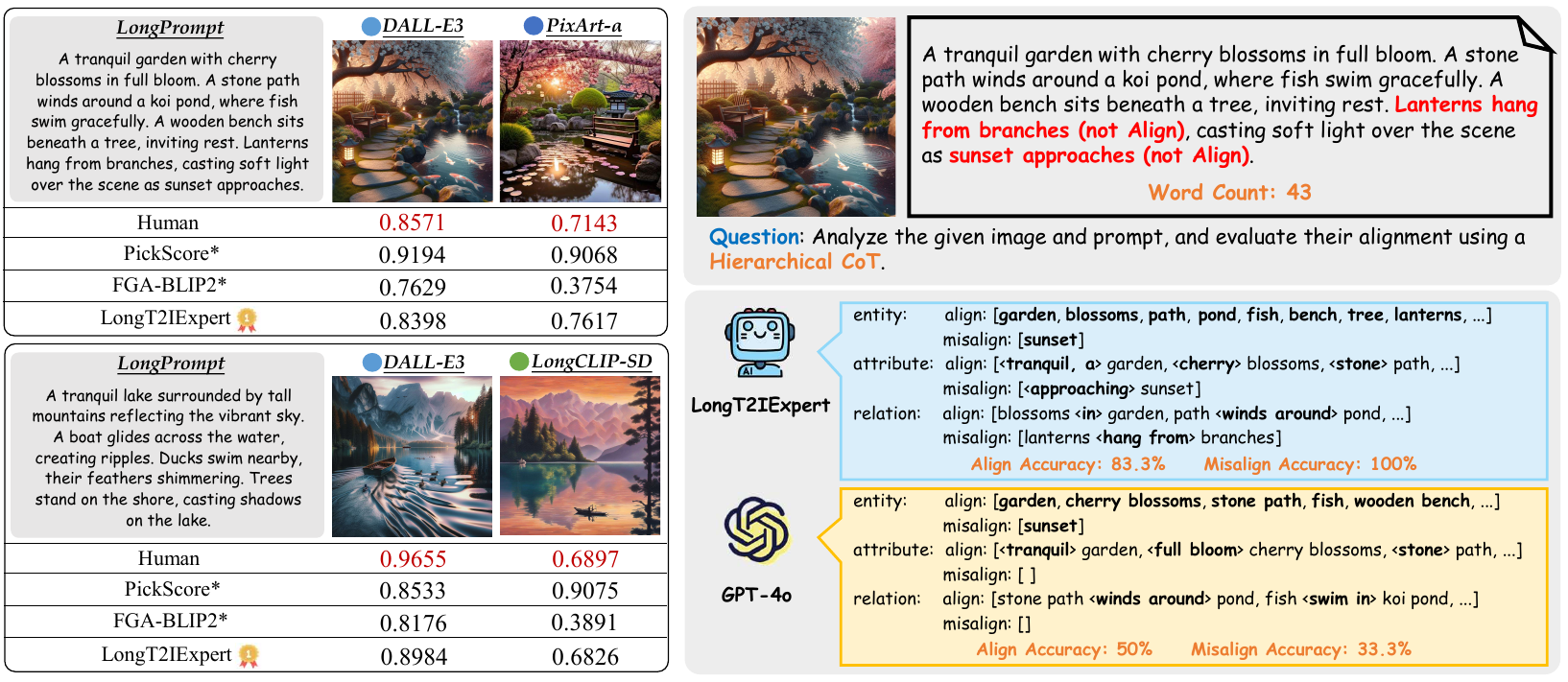}
	\caption{Visualization of long T2I alignment scoring and interpreting. Left: The predicted scores from LongT2IExpert demonstrate stronger correlation with human annotations compared to PickScore and FGA-BLIP2 (fine-tuned versions). Right: LongT2IExpert produces more accurate alignment interpretations than GPT-4o.}
	\label{fig6}
\end{figure*}

\subsection{Performance Evaluation}

\textbf{Results on Long T2I Alignment Scoring.} Table \ref{tab:1} summarizes the performance comparison between the proposed LongT2IExpert and existing T2I evaluators. To thoroughly examine these models' scoring capabilities in long-prompt scenarios, we evaluated their performance across different word-count ranges. The results demonstrate that most models show significant performance degradation as prompt length increases. Particularly in the 110+ words evaluation scenario, VLM-based models perform poorly due to text encoder's token length restrictions, while MLLM-based evaluators such as Q-Eval-Score \cite{zhang2025q} show notable performance advantages. Additionally, we fine-tuned open-source models using LongT2IBench training data for fair comparison. Our proposed LongT2IExpert demonstrates a clear advantage across all word-count intervals, suggesting its effectiveness in understanding long prompts and evaluating text-image alignment.

\textbf{Results on Long T2I Alignment Interpreting.} Due to the lack of objective alignment interpretation benchmarks and the scarcity of T2I evaluators with both scoring and interpreting capabilities, we compared LongT2IExpert with advanced MLLMs on long T2I alignment interpretation. For fair comparison, we used identical prompting patterns and output formats, including our designed HA-CoT. Table \ref{tab:2} lists the comparative results, where we report the accuracy across entity, attribute and relation. The proposed LongT2IExpert achieves the best performance, with Gemini-1.5-pro \cite{team2023gemini} ranking second. The results indicate that models generally find relation alignment more challenging than entity and attribute. In long-prompt scenarios, accurate relation judgment places higher demands on long-context understanding capabilities. We also observe that accuracy for misaligned elements is consistently lower than for aligned elements. Identifying misalignments among numerous aligned elements requires more precise localization and discrimination abilities.

\textbf{Qualitative Analysis.} In Figure \ref{fig6}-left, we present the prediction scores of LongT2IExpert, PickScore and FGA-BLIP2 (fine-tuned versions) on long text-image pairs. LongT2IExpert produces scores that demonstrate greater consistency with human annotations. For alignment interpretation, as shown in Figure \ref{fig6}-right, the proposed method generates high-accuracy alignment interpretations with superior misalignment detection capabilities. (Additional visual examples can be found in the Appendix)

\subsection{Ablation Study}
To validate the effectiveness of each component in LongT2IExpert, we conduct comprehensive ablation experiments. As summarized in Table \ref{tab:3}, we systematically evaluate the contributions of different components through various model variants. We first examine the performance when LongT2IExpert is trained solely for interpreting (w/o Score) and scoring (w/o Interpretation). The comparative results demonstrate the significant advantages of multi-task training. We then evaluate performance without the Hierarchical Alignment Chain-of-Thought (HA-CoT). The results validate the effectiveness of HA-CoT in long T2I alignment, guiding MLLMs for structured alignment reasoning.

\begin{table}[t]
	\centering
	\small
	\begin{tabular}{l|c|c}
		\toprule
		\textbf{Setting} & \textbf{Align-S} & \textbf{Align-I} \\
		\midrule
		LongT2IExpert (w/o Score) & --- & 32.8\% \\
		LongT2IExpert (w/o Interpretation) & 0.474 & --- \\
		LongT2IExpert (w/o HA-CoT) & 0.516 & 39.9\% \\
		LongT2IExpert & 0.557 & 53.2\% \\
		\bottomrule
	\end{tabular}
	\caption{Ablation study of LongT2IExpert components. The Avg. (SRCC+PLCC)/2 and Overall Acc for long T2I Alignment Scoring and Interpreting respectively, are reported.}
	\label{tab:3}
\end{table}

\section{Conclusion}
In this paper, we have performed a comprehensive subjective and objective benchmark of long text-to-image (T2I) alignment. We contribute LongT2IBench, the first-of-its-kind benchmark by introducing graph-structured annotations. We have also proposed LongT2IExpert, which fine-tunes a multimodal large language model (MLLM) to output token-level alignment scores and structured alignment interpretations. Extensive experiments demonstrate that LongT2IExpert achieves promising performance. While encouraging results have been obtained in this work, long T2I alignment remains far from solved—a challenge even for human evaluators. We hope our contributions will inspire the community to rethink the problem with more technical approaches and broader perspectives.

\section{Acknowledgments}
This work is supported by National Natural Science Foundation of China under Grants 62471349, 62171340 and 62301378,  Fundamental Research Funds for the Central Universities under Grant QTZX25076, the China Postdoctoral Science Foundation under Grant 2024M762553, and partly supported by the Innovation Fund of Xidian University under GrantYISJ25004. We also sincerely thank all annotators for their dedicated efforts and significant contributions to the construction of the benchmark.

\bibliography{aaai2026}

\end{document}